\documentclass[10pt,twocolumn]{article}

\usepackage[T1]{fontenc}
\usepackage{lmodern}
\usepackage[margin=0.72in]{geometry}
\usepackage{microtype}
\microtypesetup{expansion=false}
\usepackage{amsmath,amssymb,mathtools}
\usepackage{booktabs}
\usepackage{array}
\usepackage{multirow}
\usepackage{tabularx}
\usepackage{graphicx}
\usepackage{xcolor}
\usepackage{tikz}
\usetikzlibrary{arrows.meta,fit,positioning}
\newcounter{algorithm}
\newenvironment{algorithm}[1][t]{%
  \begin{figure}[#1]
  \refstepcounter{algorithm}
  \centering
  \hrule\smallskip
}{%
  \smallskip\hrule
  \end{figure}
}
\IfFileExists{enumitem.sty}{\usepackage{enumitem}}{}
\usepackage{url}
\providecommand{\IfDocumentMetadataT}[1]{}
\usepackage{hyperref}
\usepackage[numbers,sort&compress]{natbib}
\IfFileExists{balance.sty}{\usepackage{balance}}{\providecommand{\balance}{}}

\definecolor{minimaviolet}{HTML}{6D4AFF}
\definecolor{minimateal}{HTML}{0B9A8D}
\definecolor{minimanavy}{HTML}{15253D}
\definecolor{lightgray}{HTML}{F3F5F7}
\definecolor{warning}{HTML}{8A4B08}

\newcommand{\papertitle}{Minima-KV: Retention-Preserving KV Cache Compression with Mixed-Format Paged Attention}

\hypersetup{
  colorlinks=true,
  linkcolor=minimaviolet,
  citecolor=minimateal,
  urlcolor=minimaviolet,
  pdfauthor={Sergii Kozyrev and Davyd Maiboroda, Minima AI, Inc.},
  pdftitle={\papertitle}
}

\IfFileExists{enumitem.sty}{\setlist{nosep,leftmargin=*}}{%
  \setlength{\itemsep}{0pt}%
  \setlength{\parsep}{0pt}%
}

\newif\ifanonsubmission
\anonsubmissionfalse 

\newcommand{\system}{\textnormal{\textsc{Minima-KV}}}
\newcommand{\tqthree}{\textnormal{\textsc{TQ3}}}
\newcommand{\staletwo}{\textnormal{Stale2}}
\newcommand{\gib}{\ensuremath{\,\mathrm{GiB}}}
\newcommand{\kib}{\ensuremath{\,\mathrm{KiB}}}
\newcommand{\tps}{\ensuremath{\,\mathrm{tok/s}}}
\newcommand{\pp}{\ensuremath{\,\mathrm{pp}}}

\tikzset{
  flow/.style={-{Latex[length=1.5mm]},line width=0.55pt,color=minimanavy},
  box/.style={draw=minimanavy,rounded corners=2pt,fill=lightgray,
    minimum height=7mm,align=center,font=\footnotesize,inner sep=4pt},
  fp8box/.style={box,fill=minimateal!12},
  tqbox/.style={box,fill=minimaviolet!10},
  tnbox/.style={box,draw=warning,fill=warning!8}
}

\title{\vspace{-1.2em}\textbf{\papertitle}}

\ifanonsubmission
  \author{Anonymous Authors}
\else
  \author{
    Sergii Kozyrev \\
    Minima AI, Inc. \\
    \texttt{sergii@mnma.ai}
    \and
    Davyd Maiboroda \\
    Minima AI, Inc. \\
    \texttt{david@mnma.ai}
  }
\fi

\date{}

\begin{document}
\maketitle

\begin{abstract}
Long-context LLM serving is constrained by key--value (KV) cache capacity and
bandwidth. We present \system{}, a retention-preserving hierarchy for paged
attention. Recent and Anchor pages use FP8, while Stale pages use \tqthree{}.
Every live-request page remains addressable; format-specific kernels merge
the two physical formats without retaining a cache-sized dense shadow.

We report the available evidence as separate profiles. On Qwen3.6-27B on one
96-GB NVIDIA RTX PRO 6000 Blackwell GPU, deployment accounting reports
18.3\,KiB of attention KV per live token, a 3.50$\times$ footprint reduction
relative to BF16 and 1.75$\times$ relative to FP8, but the surviving record
does not reconcile bytes by tier. A quality-focused FP8/\tqthree{}
materializing profile matches its dense control on eight-task RULER NIAH at 16K,
improves the task-macro score by 0.20 percentage points at 8K, but regresses by
0.9 points at 4K and, on the same 503-question LongBench v2 set, by 0.80
points at 16K, 0.60 points at 32K, and 0.40 points at 64K. A separate static
direct-fused FP8/\tqthree{} canary at two
59,008-token requests measures 3.625$\times$ active-KV compression, routes all
16 full-attention layers without fallback, and retains no dense shadow. Its
single-pair throughput ratio is 0.9821 against a dense control whose KV dtype
is not bound by the surviving record. Across the three LongBench v2 lengths,
all measured deltas remain within one percentage point of dense control.
Throughput remains a separate single-pair result rather than repeated-run
evidence. Every quantitative result refers to this three-tier
Recent/Anchor/Stale system.
\end{abstract}

\section{Introduction}
\label{sec:intro}

Autoregressive decoding avoids recomputing past attention keys and values by
retaining them in a KV cache. This converts repeated computation into a state
and memory-bandwidth problem: cache size grows linearly with context length,
batch size, the number of full-attention layers, the number of KV heads, and
the head dimension. PagedAttention reduces allocator fragmentation and enables
cache sharing, but it does not reduce the information stored in each physical
page~\citep{kwon2023pagedattention}. At long context or high concurrency, KV
memory consequently limits request admission, batching, and GPU utilization.

Prior work attacks this bottleneck with quantization~\citep{liu2024kivi,
hooper2024kvquant,he2024zipcache}, eviction or sparse retention
~\citep{liu2023scissorhands,zhang2023h2o,li2024snapkv,cai2025pyramidkv}, and
low-rank or structural representations~\citep{chang2025palu,
sun2025shadowkv,krishnan2026jolt}. These directions expose a central tension.
Uniform quantization is recoverable but spends equal bits on unequal tokens.
Importance-based eviction spends memory selectively but may fail when future
attention moves to previously unimportant context. Structural compression can
preserve more information per bit, yet it requires runtime co-design to avoid
reconstruction cost overwhelming the memory saving.

\system{} treats KV state as a managed fidelity hierarchy. New pages begin in
a high-fidelity recent window. A protected anchor set retains old system
instructions, attention sinks, retrieved evidence, or other globally important
regions. Older non-anchor pages move to a three-bit representation. Every
logical page remains addressable with an exact or reconstructible
approximation; no lifecycle transition deletes KV from a live request. The
controller also supports page-level attention
scoring so repeatedly retrieved pages can move toward Anchor even when old.
The evaluated Qwen3.6 profiles in this paper use the implemented three-tier
FP8/\tqthree{} path. Completed request prefixes are a separate cache resource
and may be evicted after their last live-request reference is released.

This paper makes four contributions:

\begin{enumerate}
  \item We formulate a retention-preserving lifecycle for paged KV memory,
  implementing Recent and Anchor FP8 plus Stale \tqthree{} without deleting
  live-request pages.
  \item We describe mixed-format paged attention kernels that combine
  normalized FP8 and \tqthree{} partial outputs through a stable global
  softmax merge without constructing a cache-sized dense shadow.
  \item We report profile-separated Qwen3.6-27B evidence with measured quality
  results, including equality at 16K RULER NIAH, a within 1\% regression at 4K 
  RULER NIAH, and LongBench v2 through 64K.
  \item We provide a CUDA-graph-compatible heterogeneous decode path and an
  ownership protocol with route, fallback, dense-shadow, physical-byte, and
  immutable-prefix instrumentation.
\end{enumerate}

We do not claim the first dynamic, tiered, or mixed-precision KV method.
DynamicKV, DiffKV, HqeKV, QEvict, and MosaicKV provide close and important
precedents~\citep{zhou2025dynamickv,zhang2025diffkv,
wang2026hqekv,garg2026qevict,qiang2026mosaickv}. The evaluated
systems contribution is the integration of retention-preserving FP8/\tqthree{}
side stores, direct heterogeneous decode, ownership, and instrumentation inside
a production-oriented paged runtime.

\begin{figure*}[t]
  \centering
  \begin{tikzpicture}[node distance=7mm and 8mm]
    \node[box,minimum width=25mm] (prefill) {Prefill and\\paged allocation};
    \node[box,minimum width=27mm,right=of prefill] (controller) {Age/structure +\\optional scoring};
    \node[fp8box,minimum width=25mm,right=13mm of controller,yshift=7mm] (fp8) {Recent + Anchor\\FP8 (incl. sinks)};
    \node[tqbox,minimum width=25mm,below=3.5mm of fp8] (tq3) {Stale: packed\\\tqthree{}};
    \node[box,minimum width=28mm,right=13mm of tq3,yshift=3.5mm] (kernels) {Format-specific\\decode paths};
    \node[box,minimum width=28mm,right=of kernels] (merge) {Online-softmax\\state merge};
    \draw[flow] (prefill) -- (controller);
    \draw[flow] (controller.east) -- ++(4mm,0) |- (fp8.west);
    \draw[flow] (controller.east) -- ++(4mm,0) |- (tq3.west);
    \draw[flow] (fp8.east) -- ++(4mm,0) |- (kernels.west);
    \draw[flow] (tq3.east) -- ++(4mm,0) |- (kernels.west);
    \draw[flow] (kernels) -- (merge);
    \draw[flow,dashed,color=minimateal] (merge.north) -- ++(0,7mm) -|
      node[pos=0.55,above,font=\scriptsize] {page attention scores} (controller.north);
    \draw[flow,dashed] (tq3.north) --
      node[right,font=\scriptsize] {promote} (fp8.south);
    \node[draw=minimaviolet,dashed,rounded corners=2pt,fit=(fp8)(tq3),inner sep=3mm,
      label={[font=\scriptsize,color=minimaviolet]above:all logical pages retained}] {};
  \end{tikzpicture}
  \caption{Three-tier \system{} data path. Recent and Anchor pages use FP8,
  while Stale pages use packed \tqthree{}. The controller can consume globally
  normalized page scores; the two physical formats merge under one softmax
  normalization. All quantitative claims in this paper use this hierarchy.}
  \label{fig:overview}
\end{figure*}
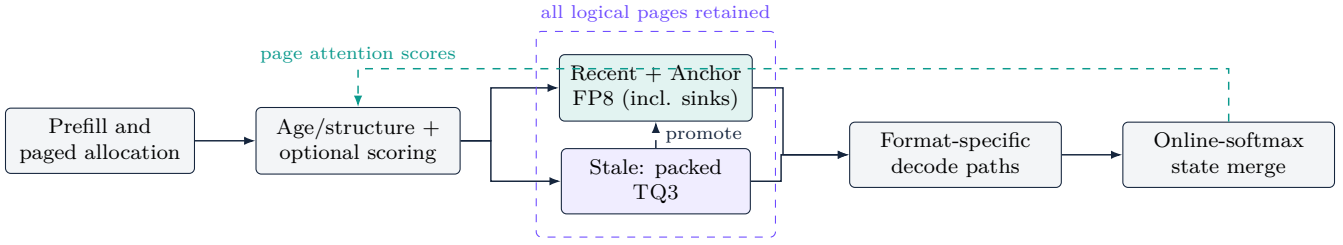

\section{Background and Motivation}
\label{sec:background}

\subsection{KV-cache scaling}

For a model with $L_f$ full-attention layers, $H_{kv}$ KV heads, head dimension
$d_h$, and $s$ bytes per scalar, the payload added by one live token is
\begin{equation}
  m_{\mathrm{KV/token}} = 2 L_f H_{kv} d_h s,
  \label{eq:kv-per-token}
\end{equation}
where the factor of two accounts for keys and values. Allocated memory also
contains scales, packed-format metadata, page padding, allocator reserve, and
temporary workspaces. In context-length labels, 1K means 1,024 tokens; binary
memory units are reported as KiB and GiB.

Qwen3.6-27B has 64 language layers arranged as 16 groups containing three
Gated DeltaNet layers followed by one full-attention layer. Its full-attention
path has four KV heads of dimension 256~\citep{qwen2026modelcard}. Therefore,
\begin{align}
 m_{\mathrm{BF16}} &= 2(16)(4)(256)(2)=65{,}536\ \mathrm{bytes/token},\\
 m_{\mathrm{FP8}}  &= 32{,}768\ \mathrm{bytes/token}.
 \label{eq:qwen-bytes}
\end{align}
The BF16 attention cache is exactly $2.00\gib$ per 32K-token sequence and
$8.00\gib$ per 128K-token sequence. The Gated DeltaNet recurrent state is a
separate per-sequence allocation; \system{} does not count it as compressed
attention KV. This distinction matters because fixed per-sequence state limits
the concurrency gain obtainable from attention-KV compression.

\subsection{Why uniform precision is suboptimal}

The next-token distribution is not equally sensitive to all past positions.
Recent tokens typically dominate local coherence and ongoing reasoning.
Older system instructions, document headings, identifiers, retrieved evidence,
and attention sinks can remain globally important. Large spans of old
background context may be less sensitive yet still become relevant after an
attention shift. A useful policy therefore distinguishes both
\emph{fidelity} and \emph{retention}: spend more bits on the first two groups,
fewer bits on Stale, and avoid irreversible deletion.

Keys and values also have different error semantics. Key distortion changes
the attention scores and can route attention to the wrong position; value
distortion changes the content read after routing. KIVI and KVQuant similarly
motivate asymmetric treatment of keys and values~\citep{liu2024kivi,
hooper2024kvquant}. \system{} preserves this separation in its codec and
kernel interfaces, even where a deployment selects the same nominal tier for
both tensors.

\section{The Minima-KV Hierarchy}
\label{sec:method}

\subsection{Paged state and lifecycle}

Let physical page $p$ contain $B_p$ logical token positions. At decode step
$t$, the controller assigns
\begin{equation}
 z_{p,t}\in\{R,A,S\},
\end{equation}
corresponding to Recent, Anchor, and Stale. A controller policy
uses page age, attention statistics, structural prompt metadata,
layer/head sensitivity, K/V role, and current memory pressure:
\begin{equation}
 z_{p,t+1}=\pi(z_{p,t},a_{p,t},h_{p,t},u_t).
 \label{eq:controller}
\end{equation}
Here $h_{p,t}$ contains structural and age features and $u_t$ is global
utilization. Attention scoring is an optional controller input. To make scores
from different physical formats comparable, page mass must be normalized by
the same global softmax used for the attention output. For query head $h$, let
$s_{h,t,i}$ be the score for token $i$ and let partition $j$ report
\begin{align}
 m_{h,t,j}&=\max_{i\in j}s_{h,t,i}, &
 \ell_{h,t,j}&=\sum_{i\in j}e^{s_{h,t,i}-m_{h,t,j}}.
 \label{eq:score-partial}
\end{align}
With $m_{h,t}=\max_j m_{h,t,j}$ and
$\ell_{h,t}=\sum_j e^{m_{h,t,j}-m_{h,t}}\ell_{h,t,j}$, define the local
mass and its global partition weight as
\begin{align}
 \rho^{(j)}_{h,t,p}
   &=\frac{1}{\ell_{h,t,j}}\sum_{i\in p}e^{s_{h,t,i}-m_{h,t,j}},\\
 w_{h,t,j}
   &=\frac{e^{m_{h,t,j}-m_{h,t}}\ell_{h,t,j}}{\ell_{h,t}}.
 \label{eq:score-global-weight}
\end{align}
Thus $q_{h,t,p}=w_{h,t,j(p)}\rho^{(j(p))}_{h,t,p}$ is globally normalized.
The reference collector conditions on page-addressable mass (excluding any
learned sink state), so it uses
$\widetilde q_{h,t,p}=q_{h,t,p}/\sum_{p'}q_{h,t,p'}$. For a sampled head set
$\mathcal H$, the controller then computes
\begin{align}
 r_{p,t} &= \frac{1}{|\mathcal H|}\sum_{h\in\mathcal H}
              \widetilde q_{h,t,p},\notag\\
 a_{p,t} &= \beta a_{p,t-1}+(1-\beta)r_{p,t}.
 \label{eq:attention-score}
\end{align}
After forced sink and Recent pages are removed from competition, a configured
policy can combine $a_{p,t}$ with age and structural features to fill the Anchor
budget. The evaluated Qwen3.6 profiles reported here had attention scoring
disabled. The checked-in reference collector computes globally normalized mass
from a retained dense view; the direct FP8/\tqthree{} kernel has not been
validated as its score source. We therefore describe scoring as a supported
mechanism and make no scoring-efficacy or scoring-overhead claim.

\paragraph{Recent ($R$, FP8).}
New KV is appended to a sliding window in FP8. This tier protects local
reasoning and avoids compression work on pages most likely to be read heavily
in the next steps.

\paragraph{Anchor ($A$, FP8).}
Anchors are protected old pages. Candidates include system and developer
instructions, initial attention sinks, retrieved evidence, document or code
structure, and pages with high cumulative or recent attention. At each retier
checkpoint, the controller ranks eligible old pages by the combined score.
Pages that leave the protected budget follow the age path toward Stale. A
Stale page is materialized and promoted when it re-enters the protected budget.

\paragraph{Stale ($S$, \tqthree{}).}
Old non-anchor pages are encoded with a three-bit rotated scalar quantizer
inspired by TurboQuant~\citep{zandieh2025turboquant}. Packed codes, scales, and required
metadata are stored next to the page. The decode kernel unpacks and applies
the representation while loading tiles, avoiding a cache-sized dequantization
buffer. The evidence records call the concrete norm-corrected packed layout
\textsc{TQ3P}; this paper uses \tqthree{} for the logical tier.

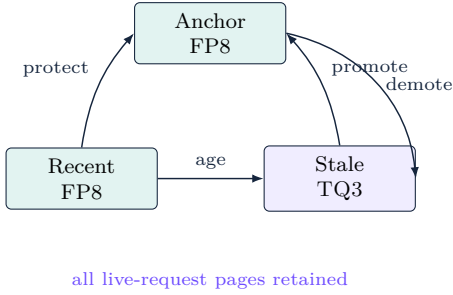
\begin{figure}[t]
  \centering
  \begin{tikzpicture}[node distance=6mm]
    \node[fp8box,minimum width=20mm] (recent) {Recent\\FP8};
    \node[tqbox,minimum width=20mm,right=14mm of recent] (stale) {Stale\\\tqthree{}};
    \node[fp8box,minimum width=20mm,above=11mm of recent,xshift=17mm] (anchor) {Anchor\\FP8};
    \draw[flow] (recent) -- node[above,font=\scriptsize] {age} (stale);
    \draw[flow] (recent.north) to[bend left=16]
      node[above left,font=\scriptsize] {protect} (anchor.west);
    \draw[flow] (stale.north) to[bend right=16]
      node[above right,font=\scriptsize] {promote} (anchor.east);
    \draw[flow] (anchor.east) to[bend left=36]
      node[right,font=\scriptsize] {demote} (stale.east);
    \node[font=\scriptsize,color=minimaviolet,below=7mm of recent,xshift=17mm]
      {all live-request pages retained};
  \end{tikzpicture}
  \caption{Implemented three-tier retention-preserving lifecycle. Recent and
  Anchor pages use FP8, while eligible old non-anchor pages use packed
  \tqthree{}. Promotion reconstructs into FP8 before publication, and every
  live-request logical page remains addressable.}
  \label{fig:lifecycle}
\end{figure}

Figure~\ref{fig:lifecycle} summarizes the implemented state transitions;
Algorithm~\ref{alg:lifecycle-protocol} specifies their copy-before-publish and
reference-release rules.

\begin{algorithm}[t]
  \label{alg:lifecycle-protocol}
  \begin{minipage}{0.96\linewidth}
  \small
  \textbf{Algorithm~\thealgorithm: Ownership-aware page-transition protocol.}
  A new mapping is published only after destination construction, and a source
  page is released only after its readers and conversion event have cleared.
  \par\smallskip\hrule\smallskip
  \begin{enumerate}
    \item \textsc{Append}$(q,p)$: allocate a request-owned Recent page, write
    KV, then publish its logical-to-physical mapping.
    \item \textsc{Retier}$(p,f)$: retain the source reference; allocate and
    populate format $f$; record conversion completion; atomically publish the
    new format tag and physical identifier; release the source only after all
    readers and the conversion event have cleared.
    \item \textsc{Promote}$(p)$: reconstruct into a newly allocated FP8 page,
    publish the FP8 mapping, then retire the compressed reference.
    \item \textsc{SealPrefix}$(p,k)$: only immutable full pages may bind to
    prefix hash $k$; the cache acquires a reference independent of requests.
    \item \textsc{Finish}$(q)$: release request references. Eviction drops the
    cache reference, and a physical page returns to its free list only after
    its last request, cache, reader, and conversion reference is gone.
  \end{enumerate}
  \end{minipage}
\end{algorithm}

\subsection{Effective rate and realized compression}

Let $p_R,p_A,p_S$ denote the fractions of logical KV scalars assigned to the
three states, with $p_R+p_A+p_S=1$. A simplified effective-rate model is
\begin{equation}
 b_{eff}=8(p_R+p_A)+b_S p_S+b_{meta},
  \label{eq:effective-bits}
\end{equation}
where $b_S$ is the realized physical \tqthree{} rate, including packed codes
and codec-local norms and scales, while $b_{meta}$ covers page indexes,
alignment, allocator effects, and remaining metadata. Compression relative to BF16 is
$C_{BF16}=16/b_{eff}$. Deployment accounting reports 18.3\,KiB per live token,
equivalent to approximately 4.58 bits per scalar, 3.497$\times$ the BF16
footprint (reported as 3.50$\times$), and 1.749$\times$ the FP8 footprint. The
aggregate does \emph{not} identify $p_R,p_A,p_S$, $b_S$, metadata, scratch, or
context-dependent occupancy. We therefore use 18.3\,KiB/token only as an
owner-reported workload aggregate and label constant-rate capacity values as
analytical scenarios rather than per-context measurements.

\subsection{Mixed-format paged attention}

For one query tile $Q$, the runtime partitions logical pages by format. Each
format-specific kernel computes a partial maximum $m_j$, exponential sum
$\ell_j$, and normalized partial output
$\bar O_j=\ell_j^{-1}\sum_{i\in j}e^{s_i-m_j}v_i$. The partitions are merged
using the stable online-softmax recurrence:
\begin{align}
 m &= \max_j m_j,\\
 \ell &= \sum_j e^{m_j-m}\ell_j,\\
 O &= \frac{\sum_j e^{m_j-m}\ell_j\bar O_j}{\ell}.
 \label{eq:softmax-merge}
\end{align}
This is algebraically equivalent to attention over the concatenated logical
cache, up to codec error and finite-precision arithmetic. In the evaluated
direct path, FP8 tiles follow the high-fidelity branch and \tqthree{} tiles
unpack while loading; page tables preserve logical ordering and the causal
mask.

The kernels are authored in TileLang~\citep{wang2025tilelang} and compiled to
CUDA. Layout, shared-memory staging, warp assignment, and reduction scheduling
are specialized by head dimension, page size, tier, and batch shape. The
performance runs report route and fallback counters so an FP8-only prompt
cannot be mistaken for evidence about the compressed path. The reported
direct-route evidence covers FP8 plus \tqthree{} only; its legacy route label
retains the \texttt{qwen32\_tq3p} kernel-family name.

\section{Implementation}
\label{sec:implementation}

\subsection{Physical stores and decode dispatch}

The evaluated Qwen3.6 path uses 1,792-token logical pages, 24 query heads,
four KV heads, and head dimension 256. The implemented physical stores in the
reported profiles are FP8 and \tqthree{}. Compact per-request page-index arrays
partition the logical block table by format while preserving order within each
set. The direct canary used two active requests, each with a 59,008-token
prompt, a 60,032-token target context, and 128 requested output tokens. Since
$59{,}008/1{,}792\approx32.93$, each request row held 33 logical block entries:
either 9 FP8 plus 24 \tqthree{}, or 10 FP8 plus 23 \tqthree{}. These entries
are logical blocks, not layers or K/V roles. The dense shadow was released.

For each stale page and KV head, \tqthree{} stores three-bit rotated key codes,
an FP16 key-norm correction, three-bit value codes, and FP16 scale/zero
metadata. The fused decode path launches branch-free FP8 and \tqthree{} scans,
reuses each unpacked KV tile across the six query heads in its GQA group, and
writes a normalized partial output plus log-sum-exp. A separate stage merges
all splits under one softmax normalization. The surviving run record does not
bind the accumulation dtype, so we make no claim about it.

The graph path is gated on a homogeneous physical-cache cohort: graph
eligibility requires the same supported Qwen3.6 geometry, live
\tqthree{} metadata, attention scoring disabled, and a released dense shadow.
Unsupported
shapes, prefill, or inconsistent cache state fail closed to the generic path.
This cohort check prevents a CUDA graph captured for a dense-retained request
from being replayed for a physically compressed request with the same tensor
shape.

\subsection{Attention-score feedback}

The checked-in reference score collector reduces globally normalized
attention from a retained dense view, stores page summaries in FP32, and
applies Eq.~\ref{eq:attention-score}. A direct mixed-format collector should
instead rescale each local page mass by Eq.~\ref{eq:score-global-weight}
before reduction. That path has not been validated: both the materializing
quality profile and the direct canary set attention scoring to off, while the
graph route also requires it to remain off. Controller behavior is therefore
outside the measured claims.

\subsection{Evaluated profiles}

\begin{table*}[t]
\centering
\caption{Evidence-bearing profiles. The rows are deliberately separate; no
single evidence profile establishes all reported properties.}
\label{tab:evaluated-policy}
\small
\begin{tabularx}{\textwidth}{@{}lXX@{}}
\toprule
Profile & Evidence boundary & Supported interpretation \\
\midrule
Qwen3.6 quality & Three-tier Recent/Anchor FP8 plus Stale \tqthree{};
1,792-token blocks; eager, batch 1;
materialization on; dynamic retiering on; scoring off; dense shadow released &
Paired RULER NIAH at 4K/8K/16K and LongBench v2 at 16K/32K/64K;
not direct-kernel correctness or speed \\
Qwen3.6 direct & Three-tier Recent/Anchor FP8 plus Stale \tqthree{};
1,792-token blocks; active 2; CUDA graph;
static page mix; dynamic retiering and scoring off; dense shadow released &
One direct-route compression/TPS canary; a separate eager quality canary using
the fast arithmetic settings fails three 16K NIAH tasks \\
Deployment accounting & Owner-reported 18.3\,KiB/token workload aggregate; no
matching per-tier reconciliation or configuration manifest & Three-tier
workload-level footprint; not a constant-rate claim across contexts \\
Qwen3.6 prefix & 16 target full-attention layers plus one MTP attention layer;
832-token blocks; Recent/Anchor FP8 plus Stale \tqthree{} & Separate Qwen3.6
warm-prefix mechanics, throughput, parity, and physical-compression validation \\
\bottomrule
\end{tabularx}
\end{table*}

Page conversion is batched by the retiering path and writes the compressed
side store before dense ownership is released. FP8-to-\tqthree{} conversion
batches eligible pages of a layer into a normalize--rotate--pack operation.
The available records do not measure conversion throughput, queue lag, peak
scratch, or the fraction hidden by serving work, so we make no asynchronous
overlap claim.

\subsection{Memory safety and prefix reuse}

Paged runtimes can retain a completed request's hashed full blocks for prefix
reuse until the physical blocks are actually reallocated. A compressed
side-store therefore needs ownership independent of any one request.
The implemented prefix path divides each format-specific pool into
live-request pages and a durable cache quota. A cache entry binds the hash of
each sealed full block to an immutable FP8 or \tqthree{} page identifier.

On a hit, the new request acquires references to those immutable pages and
installs their format tags and physical identifiers directly into its logical
block table. Compact per-format tables deduplicate pages shared by a decode
cohort; the FP8 and fused-\tqthree{} branches read the shared prefix in place
and merge their partial states under one softmax normalization.
Only the incomplete suffix is request-private: continuation allocates a new
Recent FP8 page, providing copy-on-write semantics without modifying the
cached prefix.

Reference counts span active requests, cached block hashes, and in-flight
retiering. Request completion releases only the request's references; durable
cache references keep reusable pages resident. Prefix eviction drops the cache
reference, and physical storage returns to its per-kind free list only when the
last reference and any conversion event have cleared. This ownership mechanism
was validated on Qwen3.6-27B with 16 target full-attention layers plus the MTP
attention layer, 832-token blocks, and FP8/\tqthree{} storage. This is a
separate runtime coordinate from the 1,792-token quality/direct profiles;
Appendix~\ref{app:prefix-validation} reports its numerical results.

\section{Experimental Methodology}
\label{sec:methodology}

Unless explicitly labeled analytical or external full-stack evidence, every
\system{} claim below concerns the three-tier Recent/Anchor/Stale hierarchy:
Recent and Anchor use FP8, and Stale uses \tqthree{}.

\subsection{Hardware, model, and runtime}

The Qwen3.6 experiments use one NVIDIA RTX PRO 6000 Blackwell Server Edition,
marketed with 96\,GB GDDR7 and 1,597\,GB/s memory bandwidth
~\citep{nvidia2026rtxpro6000}. The language path is Qwen3.6-27B; the vision
encoder is excluded and tensor parallelism is one. The evidence package names
the model but does not bind immutable model/tokenizer revisions, the engine
commit, or CUDA, driver, PyTorch, and TileLang versions. Those identifiers and
exact commands remain part of the publication gate in
Appendix~\ref{app:artifact-gate}.

\subsection{Baselines and causal attribution}

We distinguish theoretical storage formats from run-specific controls:

\begin{itemize}
  \item \textbf{BF16 KV}: 64\,KiB/token attention payload;
  \item \textbf{FP8 KV}: a strong production baseline at 32\,KiB/token;
  \item \textbf{three-tier \system{}}: Recent and Anchor FP8 plus Stale
  \tqthree{}, the implemented compressed policy in the Qwen3.6 quality and
  direct profiles; and
  \item \textbf{three-tier deployment aggregate}: the separately reported
  18.3\,KiB/token workload-level quantity.
\end{itemize}

The quality profile enables dynamic retiering but disables attention scoring.
The direct canary uses a static page mix and disables both. Its dense control
uses the default dense-cache policy, but the surviving record does not bind
that control's KV dtype; we therefore do not call its 0.9821 ratio a BF16 or
FP8 comparison. No unqualified ``dense'' baseline is used for a storage-ratio
claim.

The direct canary holds the model selection and serving shape fixed within its
pair, but it is not a complete baseline manifest. Full-stack deployment reports
also change weight precision and model-specific kernels. We place them in the
appendix and do not attribute their TPS or TTFT deltas to KV compression.

\subsection{Quality evaluation}

The evidence package contains two long-context evaluations:

\begin{itemize}
  \item \textbf{RULER NIAH}: eight needle-in-a-haystack tasks at 4K, 8K,
  and 16K, with 32 examples per task and length. We do not generalize this
  subset to all 13 RULER tasks~\citep{hsieh2024ruler}.
  \item \textbf{LongBench v2}: 503 multiple-choice questions across six
  realistic categories---single-document QA, multi-document QA, long
  in-context learning, dialogue-history understanding, code-repository
  understanding, and structured-data understanding
  ~\citep{bai2025longbenchv2}. The same paired question set is evaluated at
  16K, 32K, and 64K length coordinates.
\end{itemize}

Dense and compressed runs use paired examples. The conservative compressed
profile uses eager execution and materializes \tqthree{} pages before
attention. Table~\ref{tab:quality} reports exact RULER and 16K LongBench v2
scores with paired sample counts, plus the measured 32K and 64K LongBench v2
deltas on the same 503-question set. The current records do not include
paired-bootstrap intervals, a predeclared non-inferiority margin, or one fully
specified truncation rule across all three LongBench v2 lengths. The 32K
and 64K records bind aggregate paired deltas to the same materializing profile;
we do not reconstruct absolute arm counts from the rounded deltas.

\subsection{Performance evaluation}

The available performance evidence is a decode-only canary with two active
requests and 59,008 prompt tokens per request (118,016 aggregate). Both arms
record 161 counted decode tokens over 127 scheduler steps. The compressed arm
uses a static FP8/\tqthree{} mix, direct fused decode, CUDA-graph replay, no
dense shadow, and no dynamic retiering or attention scoring. This is one pair,
not a repeated end-to-end serving matrix; it has no run-to-run dispersion,
confidence interval, TTFT, or inter-token-latency percentiles. We report the
point estimate without interpreting the former 0.98--1.02 gate as statistical
equivalence.

Prefix-cache measurements come from a separate Qwen3.6-27B FP8/\tqthree{}
profile with 832-token blocks. Appendix~\ref{app:prefix-validation} reports
this warm-prefix implementation validation.

\subsection{Evidence classes}

We label configuration-linked Qwen3.6 observations \textsc{Observed--Q36},
the 18.3\,KiB/token quantity \textsc{Owner aggregate}, the Qwen3.6 warm-prefix
run \textsc{Observed--Q36-prefix}, and equation-derived capacities
\textsc{Analytical}. Vendor results that combine weights, kernels, scheduling,
and KV policy are \textsc{External full-stack}. Projected targets from the
review plan are not inserted into any result table.

\section{Results}
\label{sec:results}

\subsection{Attention-KV memory}

Table~\ref{tab:kv-memory} combines the exact architectural BF16/FP8 rates with
the owner-reported 18.3\,KiB/token aggregate. Because the latter lacks a
per-context tier reconciliation, every context row is a constant-rate
analytical projection, not an additional measurement. At one million live
tokens, FP8 and the 18.3-KiB scenario require 30.52 and 17.45\,GiB. The
aggregate uses 71.4\% fewer bytes than BF16 and 42.8\% fewer than FP8.

\begin{table}[t]
\centering
\caption{Analytical attention-KV payload by live context. BF16 and FP8 follow
Eq.~\ref{eq:kv-per-token}; the final column applies the owner-reported
18.3\,KiB/token aggregate uniformly. Allocator reserve, metadata not included
in that aggregate, scratch, and recurrent state are excluded.}
\label{tab:kv-memory}
\small
\begin{tabular}{@{}lrrr@{}}
\toprule
Live tokens & BF16 & FP8 & \system{} \\
\midrule
8K      & 0.500\gib & 0.250\gib & 0.143\gib \\
32K     & 2.000\gib & 1.000\gib & 0.572\gib \\
128K    & 8.000\gib & 4.000\gib & 2.288\gib \\
262,144 & 16.000\gib & 8.000\gib & 4.575\gib \\
1,000,000 & 61.04\gib & 30.52\gib & 17.45\gib \\
\midrule
Bytes/token & 64.0\kib & 32.0\kib & 18.3\kib \\
BF16 compression & 1.00$\times$ & 2.00$\times$ & 3.497$\times$ \\
\bottomrule
\end{tabular}
\end{table}

\subsection{Long-context quality}

The materializing FP8/\tqthree{} quality profile disables attention scoring.
It is not a broad parity pass: the exact paired results show equality at 16K
RULER NIAH, a small positive 8K task-macro delta, and a regression at 4K.
On LongBench v2, the regression narrows from 0.80 percentage points at 16K to
0.60 points at 32K and 0.40 points at 64K.

\begin{table}[t]
\centering
\caption{Paired quality results for the materializing FP8/\tqthree{} profile.
RULER values are task-macro scores across eight NIAH tasks with 32 examples
per task. LongBench v2 uses the same 503 paired questions at each length; the
16K row reports exact arm counts, while the 32K and 64K rows report measured
paired deltas. No confidence intervals or predeclared parity margins are
available.}
\label{tab:quality}
\footnotesize
\setlength{\tabcolsep}{2.2pt}
\begin{tabular}{@{}lrrrr@{}}
\toprule
Evaluation & $n$ & Control & Minima & Delta \\
\midrule
NIAH 4K & 256 & 97.27\% & 93.75\% & $-0.9\pp$ \\
NIAH 8K & 256 & 99.80\% & 100.00\% & $+0.20\pp$ \\
NIAH 16K & 256 & 100.00\% & 100.00\% & $0.00\pp$ \\
LBv2 cap 16K & 503 & 190 (37.77\%) & 186 (36.98\%) & $-0.80\pp$ \\
LBv2 32K & 503 & \multicolumn{2}{c}{paired delta record} & $-0.60\pp$ \\
LBv2 64K & 503 & \multicolumn{2}{c}{paired delta record} & $-0.40\pp$ \\
\bottomrule
\end{tabular}
\end{table}

At 4K, the aggregate corresponds to 249/256 effective correct for the dense
control and 240/256 for \system{}. The losses concentrate in
\texttt{niah\_multikey\_3} ($0.8125\rightarrow0.5625$) and
\texttt{niah\_multiquery} ($0.96875\rightarrow0.9375$). On capped-16K
LongBench v2, six predictions change: five dense-correct/Minima-wrong cases
and one gain. The 32K and 64K rows extend the same profile and paired
503-question set; their records bind the measured aggregate deltas without
reconstructing absolute arm counts.
These observations motivate paired intervals and category-stratified reporting
in the final experiment rather than a qualitative ``no regression'' verdict.

The corresponding processes peak at 92,024 versus 60,704\,MiB for NIAH 16K
and 92,076 versus 61,290\,MiB for capped-16K LongBench v2. Those 31-GiB
differences are
not active-request KV measurements: one 16K BF16 full-attention cache is only
about 1\,GiB, and the runs use differently sized preallocated KV arenas plus
weights, recurrent state, workspaces, and allocator reserve. We therefore do
not use process peak HBM to estimate codec compression.

\subsection{Fused-decode performance and quality canary}

The active-two canary exercises two 59,008-token prompts through the direct
FP8/\tqthree{} route. Its page mix is static; dynamic retiering and attention
scoring are disabled. Both arms count 161 decode tokens over 127 scheduler
steps. Dense-control wall time is 5.401996\,s and compressed wall time is
5.500437\,s, yielding 29.804 and 29.270 counted tok/s. This reconciles the
throughput values with the 42.535 and 43.311\,ms scheduler-step averages:
active batch two does not imply that every measured scheduler step contributes
two counted output tokens.

The route label is
\path{custom_mixed_qwen32_tq3p_cuda_graph}. Instrumentation records two
request-layer instances for each of Qwen3.6's 16 physical full-attention
layers, so the 32-count denominator is not a 32-layer model claim. The direct
route covers 16/16 physical attention layers, with 0/32 request-layer
fallbacks and no dense shadow. The control's KV dtype is not bound by the
surviving record; 0.9821 is therefore reported only as a within-pair ratio.
The active-cache counter prices 131,610,624 compressed bytes against
477,102,080 full-cache bytes (3.6251$\times$); the broader accounting ratio is
3.6146$\times$. This profile and accounting window differ from the
18.3\,KiB/token deployment aggregate, and no shared occupancy manifest exists
to reconcile the two rates.

\begin{table}[t]
\centering
\caption{Single-pair direct FP8/\tqthree{} decode canary, active batch two and
59,008 prompt tokens per request. The dense-control KV dtype is not bound.}
\label{tab:kernel-canary}
\footnotesize
\setlength{\tabcolsep}{2pt}
\begin{tabular}{@{}lrr@{}}
\toprule
Metric & Control/ref. & Direct Minima \\
\midrule
Decode throughput & 29.804\tps & 29.270\tps \\
Step time & 42.535\,ms & 43.311\,ms \\
Tokens / sched. steps & 161 / 127 & 161 / 127 \\
Relative throughput & 1.000 & 0.9821 \\
Physical compression & 1.00$\times$ & 3.625$\times$ \\
Active KV bytes & 477,102,080 & 131,610,624 \\
Layer routes & --- & 16/16 \\
Fallbacks (request-layer) & --- & 0/32 \\
\bottomrule
\end{tabular}
\end{table}

The fast arithmetic settings do \emph{not} clear their separate 16K RULER NIAH
quality canary. With 16 examples per task, scores fall from 1.0 to 0.875 on
\texttt{niah\_multikey\_1}, 0.828125 on \texttt{niah\_multiquery}, and
0.78125 on \texttt{niah\_multivalue}; the other five tasks remain at 1.0.
That quality run uses eager execution, so the failure is below CUDA-graph
replay and prevents combining the speed and quality claims into one profile.

\subsection{Concurrency capacity}

Raw codec ratio is not request capacity. Let $M$ be the serving memory
envelope, $W$ resident weights and shared runtime state, $D$ the fixed
per-sequence non-attention state, and $K(L)$ attention KV for a sequence of
length $L$. The memory ceiling is
\begin{equation}
 C_{mem}(L)=\left\lfloor\frac{M-W}{D+K(L)}\right\rfloor.
 \label{eq:capacity}
\end{equation}
We retain $M=86.4\gib$, $W=25.15\gib$, and $D=0.153\gib$ only as the historical
analytical scenario used in the deployment report. It is not derived from the
vendor's decimal ``96 GB'' label. A separate CUDA-visible probe reports
101,975,851,008 bytes (94.972\,GiB), for which 90\% would be 85.475\,GiB;
substituting that value would require refitting $D$ rather than mixing the two
models. Holding the historical scenario fixed and changing only KV rate yields
Table~\ref{tab:capacity}. At 32K, FP8 supports 53 resident contexts and an exact
3.50$\times$-BF16 KV scenario supports 84, a 1.58$\times$ analytical gain over
FP8.

\begin{table}[t]
\centering
\caption{Derived resident-context memory ceilings under Eq.~\ref{eq:capacity}
with historical scenario parameters $M=86.4\gib$, $D=0.153\gib$, and
$W=25.15\gib$. Every cell is analytical; the Minima column assumes a constant
exact 3.50$\times$ BF16 rate. These are not simultaneous-decode SLO guarantees.}
\label{tab:capacity}
\small
\setlength{\tabcolsep}{3.2pt}
\begin{tabular}{@{}lrrr@{}}
\toprule
Context & BF16 KV & FP8 KV & Minima KV \\
\midrule
8K   & 93 & 151 & 207 \\
16K  & 53 & 93 & 139 \\
32K  & 28 & 53 & 84 \\
64K  & 14 & 28 & 47 \\
128K & 7 & 14 & 25 \\
262K & 3 & 7 & 12 \\
\bottomrule
\end{tabular}
\end{table}

Resident capacity and active decode concurrency are different. Compute,
memory bandwidth, scheduler policy, and latency objectives can bind before
HBM. The separate full-stack report observes 98 admitted 32K context
equivalents, whereas Eq.~\ref{eq:capacity} with its compressed-weight input
predicts 97; Appendix~\ref{app:external} keeps the observed and analytical
quantities separate. Neither number claims 98 simultaneous decodes at unchanged
inter-token latency.

\subsection{Service-level implications}

At low load and fixed weights, post-prefill KV compression is not expected to
improve compute-only TTFT; conversion can add cost. At
high load, smaller resident state can reduce admission blocking and scheduler
queueing. Service-observed TTFT can therefore improve even when prefill kernel
latency is unchanged:
\begin{equation}
 T_{TTFT}=T_{queue}+T_{prefill}+T_{schedule}.
 \label{eq:ttft}
\end{equation}
The available data do not include an open-loop offered-load experiment, so we
make no KV-only p95/p99 tail-latency claim. Such a claim requires an offered-load
sweep that reports admission, queueing, and tail latency separately.

\section{Related Work}
\label{sec:related}

\paragraph{Low-bit KV quantization.}
KIVI quantizes keys per channel and values per token at two bits
~\citep{liu2024kivi}. KVQuant combines pre-RoPE key quantization, nonuniform
formats, and dense-and-sparse outlier handling~\citep{hooper2024kvquant}.
ZipCache assigns precision using normalized attention saliency
~\citep{he2024zipcache}; MiKV retains low-importance tokens at lower precision
rather than deleting them~\citep{yang2024mikv}. TurboQuant uses random
rotation and scalar quantization to obtain near-optimal online distortion
~\citep{zandieh2025turboquant}. These methods establish both low-bit KV and
mixed fidelity as prior art. \system{} differs in coupling low-bit pages to a
retained FP8/\tqthree{} lifecycle and direct mixed-format decode inside a paged
serving runtime.

\paragraph{Eviction and sparse retention.}
Scissorhands exploits persistence of token importance
~\citep{liu2023scissorhands}; H2O keeps recent tokens and cumulative
heavy-hitters~\citep{zhang2023h2o}; SnapKV infers important prompt positions
from an observation window~\citep{li2024snapkv}; PyramidKV varies budgets by
layer~\citep{cai2025pyramidkv}. These methods can produce large memory and
compute savings, but removed KV is unavailable after an attention shift.
\system{} instead maintains an approximate representation for every retained
logical page.

\paragraph{Dynamic and hybrid management.}
DynamicKV reallocates cache budget across layers and tasks
~\citep{zhou2025dynamickv}. DiffKV differentiates keys and values and combines
dynamic sparsity with paging and compaction~\citep{zhang2025diffkv}. HqeKV
uses multiple precision levels plus eviction and periodic reclassification
~\citep{wang2026hqekv}. QEvict exposes full, quantized, and deleted states with
promotion from its quantized tier~\citep{garg2026qevict}. MosaicKV selects
sequence- and channel-dimension strategies per segment and supplies packed
kernels~\citep{qiang2026mosaickv}. RDKV casts quantization and eviction as
endpoints of a rate--distortion bit-allocation problem~\citep{zhang2026rdkv}.
Accordingly, our contribution is not dynamic tiering in isolation; the
evaluated contribution is retention-preserving FP8/\tqthree{} page ownership
plus direct mixed-format implementation.

\paragraph{Head- and layer-aware policies.}
DuoAttention separates retrieval heads, which retain full attention, from
streaming heads with bounded caches~\citep{xiao2024duoattention}; PyramidKV
and DynamicKV vary budgets across layers or tasks. These policies are
complementary to \system{}'s page-format controller. The optional controller
aggregates globally normalized attention mass into an EWMA page score and can
use it to retain or promote FP8 anchors without changing the physical codec.
Scoring is disabled in the reported Qwen3.6 profiles.

\paragraph{Structural compression and offload.}
Palu caches low-rank latent states derived from decomposed projection weights
~\citep{chang2025palu}. ShadowKV retains low-rank pre-RoPE keys on GPU and
offloads values, reconstructing a selected sparse subset
~\citep{sun2025shadowkv}. JoLT combines Tucker compression with a rotated
residual~\citep{krishnan2026jolt}. These works make any ``first tensor
decomposition'' claim untenable. GEAR combines low-bit quantization with
low-rank and sparse residual corrections~\citep{kang2024gear}. These structural
approaches are distinct from the three-tier FP8/\tqthree{} system evaluated
here.

\begin{table*}[t]
\centering
\caption{Closest literature-reported systems results. Values use each paper's
own model, hardware, baseline, and protocol and are not an apples-to-apples
ranking.}
\label{tab:related-comparison}
\small
\begin{tabularx}{\textwidth}{@{}lXXX@{}}
\toprule
Method & Mechanism & Retention semantics & Literature-reported result \\
\midrule
RDKV~\citep{zhang2026rdkv} & Joint 0-to-full-bit rate--distortion allocation
over tokens/channels & Zero bits permits eviction & 97.81\% of full-cache
LongBench accuracy at 2.48\% retention; 4.5$\times$ decode speedup and
1.9$\times$ peak-memory reduction at 128K \\
MosaicKV~\citep{qiang2026mosaickv} & Dynamic sequence- and channel-dimension
compression by segment & Different compressed-management semantics &
3$\times$ memory reduction, 1.76\% average accuracy loss, and up to
7.3$\times$ throughput versus its uncompressed baseline \\
JoLT~\citep{krishnan2026jolt} & Partial Tucker plus rotated low-bit residual &
Compressed approximation without reported token eviction & 2--3$\times$
near-lossless compression; compression-time rather than serving-throughput
speedup \\
\system{} (this paper) & Retained Recent/Anchor FP8 and Stale \tqthree{} pages
with direct mixed decode & Every live-request logical page remains addressable &
18.3\,KiB/token owner aggregate; one direct canary at 0.9821$\times$ its dense
control; quality profile has measured regressions \\
\bottomrule
\end{tabularx}
\end{table*}

LMCache and Mooncake build placement hierarchies across GPU memory, host DRAM,
storage, and network resources~\citep{cheng2025lmcache,qin2024mooncake}.
\system{}'s tiers are on-GPU fidelity states rather than placement states, so
the two approaches can compose: an FP8 or \tqthree{} page can still be moved by
an external cache layer.

\section{Limitations}
\label{sec:limitations}

{\small
\begin{itemize}
  \item \textbf{One model and GPU.} Results are for Qwen3.6-27B on one RTX PRO
  6000; other accelerators and non-hybrid models may differ. Tensor-parallel,
  multi-GPU, and vision paths are not evaluated.
  \item \textbf{Hybrid-model denominator.} Only 16 of 64 language layers hold
  full-attention KV; the 18.3\,KiB/token aggregate applies to that pool, not
  Gated DeltaNet state or total process HBM.
  \item \textbf{Unreconciled aggregate.} The 18.3\,KiB/token quantity lacks
  per-tier bytes, metadata, scratch, dense-shadow accounting, context-specific
  occupancy distributions, and a matching immutable manifest, so it does not
  establish a constant compression rate across lengths.
  \item \textbf{Quality coverage.} The materializing profile regresses 0.90 at 4K
  RULER NIAH and by 0.80, 0.60, and 0.40 percentage points on LongBench v2 at
  16K, 32K, and 64K. The 32K/64K records report aggregate paired deltas rather than
  absolute arm counts, and no paired confidence interval is available.
  \item \textbf{Profile separation.} Quality, direct decode, deployment
  accounting, and prefix reuse come from different configurations. The direct
  canary is one pair, its dense-control KV dtype is unbound, and it has no
  repeated-run uncertainty or latency distribution.
  \item \textbf{Unevaluated mechanisms.} Attention scoring is off in both
  Qwen3.6 profiles. Conversion throughput, scratch, and asynchronous overlap
  are also unmeasured.
  \item \textbf{Full-stack confounding.} Public TPS and TTFT improvements also
  use NVFP4 weights and model-specific kernels; a fixed-weight factorial
  ablation is not available.
  \item \textbf{Resident versus active concurrency.} Higher admission capacity
  does not guarantee unchanged latency at equal active concurrency.
\end{itemize}
}

\section[Future Directions and Conclusion]{Future Directions and\\Conclusion}
\label{sec:future}

We presented \system{}, a retention-preserving three-tier paged-KV runtime:
Recent and Anchor pages use FP8, while Stale pages use \tqthree{}. It retains
every live-request page through ownership-safe format changes, executes both
physical formats through a globally normalized online-softmax merge, releases
the cache-sized dense shadow on the physical-compression path, and reports each
evidence profile within its bound configuration.

On Qwen3.6-27B, deployment accounting reports 18.3\,KiB/token, equivalent to
3.497$\times$ compression relative to BF16 and 1.749$\times$ relative to FP8,
with per-tier reconciliation remaining part of the publication gate. The
materializing profile matches its dense control at 16K RULER NIAH and regresses
at 4K. On LongBench v2, its measured deltas are $-0.80\pp$ at 16K,
$-0.60\pp$ at 32K, and $-0.40\pp$ at 64K. A separate direct canary achieves
3.625$\times$ active-KV compression and 0.9821$\times$ within-pair throughput
while routing all 16 full-attention layers with no dense shadow.

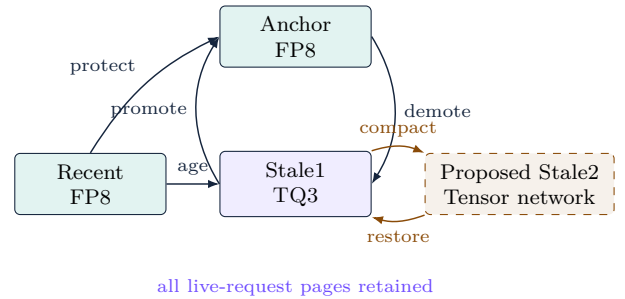
\begin{figure}[!tbp]
  \centering
  \begin{tikzpicture}[node distance=6mm]
    \node[fp8box,minimum width=20mm] (future-recent) {Recent\\FP8};
    \node[tqbox,minimum width=20mm,right=7mm of future-recent]
      (future-stale1) {Stale1\\\tqthree{}};
    \node[tnbox,dashed,minimum width=25mm,right=7mm of future-stale1]
      (future-stale2) {Proposed \staletwo{}\\Tensor network};
    \node[fp8box,minimum width=20mm,above=11mm of future-stale1]
      (future-anchor) {Anchor\\FP8};
    \draw[flow] (future-recent) -- node[above,font=\scriptsize] {age} (future-stale1);
    \draw[flow] (future-recent.north) to[bend left=16]
      node[above left,font=\scriptsize] {protect} (future-anchor.west);
    \draw[flow] (future-stale1.west) to[bend left=32]
      node[left,font=\scriptsize] {promote} (future-anchor.west);
    \draw[flow] (future-anchor.east) to[bend left=32]
      node[right,font=\scriptsize] {demote} (future-stale1.east);
    \draw[flow,color=warning] (future-stale1.north east) to[bend left=20]
      node[above,font=\scriptsize] {compact} (future-stale2.north west);
    \draw[flow,color=warning] (future-stale2.south west) to[bend left=18]
      node[below,font=\scriptsize] {restore} (future-stale1.south east);
    \node[font=\scriptsize,color=minimaviolet,below=7mm of future-stale1]
      {all live-request pages retained};
  \end{tikzpicture}
  \caption{Future four-tier lifecycle. The measured three-tier system would
  relabel Stale as Stale1 and add proposed \staletwo{} tensor-network groups
  with compact and restore transitions.}
  \label{fig:lifecycle-future}
\end{figure}

\paragraph{Toward a fourth tier with a shared tensor backbone.}
A future hierarchy could extend the implemented state set
$\{R,A,S\}$ to $\{R,A,S_1,S_2\}$: the current Stale state would become
Stale1, and the coldest retained page groups would enter a proposed \staletwo{}
tensor-network representation. Fig.~\ref{fig:lifecycle-future} illustrates
this prospective lifecycle. Every quantitative claim in this paper refers to
the three-tier Recent/Anchor/Stale system above.

Our earlier Minima model-compression work used sensitivity-guided Tucker,
tensor-train, and tensor-ring factorizations and proposed a shared tensor
backbone with per-layer adapters~\citep{kozyrev2026tensor}. \staletwo{} is the
cache-side analogue---not a measured result. It would amortize a common Tucker
or tensor-train core across compatible full-attention layers and cold page
groups, with key/value- and group-specific adapters. Tiles would reconstruct
on demand, follow the existing copy-before-publish protocol, and emit
$(m_j,\ell_j,\bar O_j)$ for the global-softmax merge~\citep{krishnan2026jolt}.

Testing this direction requires a frozen configuration binding ranks,
residuals, factor and metadata bytes, conversion and scratch costs, and the
tile-decode route to joint quality, throughput, and prefix-correctness
evaluation. Only the Recent/Anchor/Stale FP8/\tqthree{} runtime is measured
here.

\clearpage
\onecolumn
\appendix

\section{Artifact Index and Publication Gate}
\label{app:artifact-gate}

Table~\ref{tab:artifact-status} summarizes the evidence classes used by this
paper and the scope assigned to each. No public artifact URL is assigned yet;
the paper should not be submitted until the bundle below is published.

\begin{center}
\centering
\refstepcounter{table}\label{tab:artifact-status}
\textbf{Table~\thetable: Configuration-linked evidence used in this paper.}
\par\medskip
\footnotesize
\begin{tabularx}{\textwidth}{@{}lX@{}}
\toprule
Evidence & Scope \\
\midrule
Qwen3.6 materializing quality &
Three-tier Recent/Anchor FP8 plus Stale \tqthree{}, scoring off; exact
4K/8K/16K NIAH and capped-16K LongBench v2 scores; measured 32K/64K
LongBench v2 deltas on the same 503-question set \\
Qwen3.6 direct decode &
Static three-tier FP8/\tqthree{} route, active 2, physical bytes, TPS,
dense-shadow and fallback counters \\
Direct control &
Within-pair TPS control; KV dtype not bound by the surviving summary \\
Fast-profile quality &
Eager direct arithmetic canary; three of eight NIAH tasks regress \\
Qwen3.6 prefix validation &
16 target attention layers plus one MTP attention layer; 832-token pages;
Recent/Anchor FP8 plus Stale \tqthree{}; N=32 warm-prefix coordinate \\
\bottomrule
\end{tabularx}
\end{center}

The implementation is proprietary and is not released with this paper. 
We report the model and tokenizer versions, hardware and software stack, 
benchmark protocols, and configuration-linked aggregate results needed to 
interpret the evaluation. Detailed manifests, telemetry, and raw logs are 
retained internally for audit but are not part of a public release.

\section{Claim Ledger}
\label{app:claims}

\begin{center}
\centering
\refstepcounter{table}\label{tab:claim-ledger}
\textbf{Table~\thetable: Evidence classification for quantitative claims.}
\par\medskip
\small
\begin{tabularx}{\textwidth}{@{}lXX@{}}
\toprule
Class & Evidence & Supported statement \\
\midrule
Owner aggregate & 64.0, 32.0, and 18.3\,KiB/token & 18.3 implies
3.497$\times$ BF16 and 1.749$\times$ FP8 compression; tier attribution and
context distribution are unproven. \\
Observed--Q36 & Materializing quality profile & 4K NIAH $-0.9\pp$; 8K
$+0.20\pp$; 16K equality; LongBench v2 $-4/503$ ($-0.80\pp$) at 16K,
$-0.60\pp$ at 32K, and $-0.40\pp$ at 64K. \\
Observed--Q36 & Single direct canary & 3.625$\times$ active-KV compression,
0.9821$\times$ within-pair TPS, 16/16 physical layers routed, 0/32
request-layer fallbacks, and no dense shadow; control KV dtype unbound. \\
Observed--Q36 & Fast arithmetic quality canary & Three 16K NIAH tasks within 1\% of dense control.
the direct speed and broad quality claims cannot be combined. \\
Observed--Q36-prefix & Qwen3.6 warm-prefix run & 1.021$\times$ dense-FP8 TPS,
32/32 exact outputs, 4.342$\times$ physical KV compression, and zero fallbacks
at the reported N=32 coordinate. \\
External full-stack & Rafay/Qdrant reports & Deployment outcomes combine
weight, kernel, scheduler, and KV changes and are not KV-only causal estimates. \\
Analytical & Eq.~\ref{eq:capacity} & Historical scenario predicts 53 FP8 and
84 exact-3.50$\times$ Minima resident 32K contexts; these are not active-decode
SLO measurements. \\
Proposed & Direct mixed-format attention-score collection & Three-tier
controller design only; excluded from measured claims. \\
\bottomrule
\end{tabularx}
\end{center}

\section{Qwen3.6 Warm-Prefix Validation}
\label{app:prefix-validation}

The warm-prefix record uses Qwen3.6-27B with 16 target full-attention layers
plus the MTP attention layer, 832-token blocks, and FP8/\tqthree{} storage. The
coordinate is N=32, 32,256 prompt tokens, 80 requested output tokens,
replay-one, and batched bound-prefix prefill. It is distinct from the
1,792-token Qwen3.6 quality/direct coordinates.

\begin{center}
\centering
\refstepcounter{table}\label{tab:prefix-cache}
\textbf{Table~\thetable: Qwen3.6 warm-prefix results against dense FP8
automatic prefix caching (APC).}\par\medskip
\small
\begin{tabular}{@{}lr@{}}
\toprule
Metric & Result \\
\midrule
Warm-prefix throughput / dense FP8 APC & 1.021$\times$ \\
Steady throughput, Minima / dense & 317.69 / 311.28\tps \\
Exact passkey outputs & 32/32 \\
Duplicate-batch token parity & true \\
Physical KV compression & 4.342$\times$ \\
Durable compressed pages & 37 \\
TileLang uses / fallbacks & 1,632 / 0 \\
\bottomrule
\end{tabular}
\end{center}

The final r14 run uses the shared-page TileLang cohort for three-row MTP calls;
the one-row cohort mode remains opt-in. The 4.342$\times$ value is the physical
KV compression counter for this warm-prefix coordinate, while 1.021$\times$
is its paired steady-throughput ratio.

\section{External Full-Stack Deployment Reports}
\label{app:external}

Two Minima AI reports provide deployment context but are not causal evaluations
of KV compression. Rafay compares FP8 weights and FP8 KV with an NVFP4 W4A4
stack that also enables model-specific kernels and \system{}
~\citep{minima2026rafay}. Qdrant compares a BF16 full stack with a full Minima
stack~\citep{minima2026qdrant}.

\begin{center}
\centering
\refstepcounter{table}\label{tab:public-results}
\textbf{Table~\thetable: Self-published full-stack deployment evidence.
Weight, kernel, scheduler, and KV changes are combined.}\par\medskip
\small
\begin{tabular}{@{}llrrr@{}}
\toprule
Study & Metric & Baseline & Full Minima & Change \\
\midrule
Rafay (FP8 baseline) & Resident weight payload & 25.15\gib & 15.72\gib & $-37.5\%$ \\
 & Attention KV at 1M live tokens & 30.52\gib & 17.45\gib & $-42.8\%$ \\
 & Resident 32K context equivalents & 53 & 98 & $1.85\times$ \\
 & Aggregate output throughput & 254.0\tps & 387.6\tps & $1.53\times$ \\
 & TTFT p50 / p95 & 1.21 / 2.29\,s & 0.91 / 1.73\,s & $-24.8/-24.5\%$ \\
\midrule
Qdrant (BF16 baseline) & Nominal weights & 50.3\gib & 15.7\gib & $3.20\times$ compression \\
 & Attention KV / active token & 64.0\kib & 18.3\kib & $3.50\times$ compression \\
 & Admitted 32K sessions & 11 & 96 & $8.7\times$ \\
 & Standalone output throughput & 206.4\tps & 392.2\tps & $1.90\times$ \\
 & Agent latency p50 / p95 & 14.6 / 20.8\,s & 7.7 / 11.0\,s & $-47.3/-47.1\%$ \\
\bottomrule
\end{tabular}
\end{center}

With $W=15.72\gib$, the historical capacity equation predicts 97 rather than
the Rafay report's observed 98 contexts, showing that its fitted envelope is
only approximate. The Qdrant BF16 row admits 11 sessions, whereas substituting
its nominal weights into the Rafay envelope predicts roughly 17. The studies
do not expose matching memory envelopes and block reservations, so they are
incomparable operating points.

\clearpage
\twocolumn
\balance

\end{document}